\documentclass[sigconf]{acmart}
\usepackage{algorithm}
\usepackage{booktabs}
\usepackage{graphicx}
\usepackage{amsmath}
\usepackage{comment}
\usepackage{amsfonts}
\usepackage{xcolor}
\usepackage{multirow}
\usepackage{verbatim}
\usepackage{multicol}
\usepackage{subfigure}
\usepackage{extarrows}
\usepackage{algorithm}  
\usepackage{algorithmicx}  
\usepackage{algpseudocode}  
\usepackage{amsmath} 
\usepackage{color}
\definecolor{ourdarkgreen}{RGB}{84,130,53}
\definecolor{ourdarkblue}{RGB}{68,114,195}
\definecolor{}{RGB}{255,240,245}

\usepackage{listings}
\usepackage{color}

\definecolor{dkgreen}{rgb}{0,0.6,0}
\definecolor{gray}{rgb}{0.5,0.5,0.5}
\definecolor{mauve}{rgb}{0.58,0,0.82}

\AtBeginDocument{%
  \providecommand\BibTeX{{%
    \normalfont B\kern-0.5em{\scshape i\kern-0.25em b}\kern-0.8em\TeX}}}

\copyrightyear{2021} 
\acmYear{2021} 
\setcopyright{acmcopyright}\acmConference[IJCKG'21]{The 10th International Joint Conference on Knowledge Graphs}{December 6--8, 2021}{Virtual Event, Thailand}
\acmBooktitle{The 10th International Joint Conference on Knowledge Graphs (IJCKG'21), December 6--8, 2021, Virtual Event, Thailand}
\acmPrice{15.00}
\acmDOI{10.1145/3502223.3502237}
\acmISBN{978-1-4503-9565-6/21/12}

\begin{document}
%\settopmatter{printacmref=false} \setcopyright{none}
%\renewcommand \footnotetextcopyrightpermission[1]{} \pagestyle{plain}
 
\title{Normal vs. Adversarial:
Salience-based Analysis of Adversarial Samples for Relation Extraction}


\author{
Luoqiu Li$^{1,2*}$, 
Xiang Chen$^{4*}$, 
Zhen Bi$^{1,2*}$, 
Xin Xie$^{1,2*}$, 
Shumin Deng$^{1,2}$, 
Ningyu Zhang$^{1,2\star}$, \\
Chuanqi Tan$^{3}$, 
Mosha Chen$^{3}$, 
Huajun Chen$^{1,2\star}$
}
% \author{Ningyu Zhang$^{1,2}$*, Qianghuai Jia$^{3}$*, Shumin Deng$^{1,2}$*, Xiang Chen$^{1,2}$, Hongbin Ye$^{1,2}$, Hui Chen$^{3}$, Huaixiao Tou$^{3}$, Gang Huang$^{4}$, Zhao Wang$^{1}$,  Nengwei Hua$^{3}$, Huajun Chen$^{1,2}\dagger$}

\affiliation{
$^1$Zhejiang University\country{China} \& AZFT Joint Lab for Knowledge Engine\country{China}
$^3$Alibaba Group\country{China}
}
\affiliation{
$^2$Hangzhou Innovation Center\country{China}, Zhejiang University\country{China},
}

\email{
{luoqiu.li,xiang_chen,bizhen_zju,xx2020,231sm,zhangningyu,huajunsir}@zju.edu.cn, 
}
\email{
{chuanqi.tcq,chenmosha.cms}@alibaba-inc.com
}
\renewcommand{\shortauthors}{Luoqiu Li, et al.}

 
\begin{abstract}
Recent neural-based relation extraction approaches, though achieving promising improvement on benchmark datasets, have reported their vulnerability towards adversarial attacks. Thus far, efforts mostly focused on generating adversarial samples or defending adversarial attacks, but little is known about the difference between normal and adversarial samples. In this work, we take the first step to leverage the salience-based method to analyze those adversarial samples. We observe that salience tokens have a direct correlation with adversarial perturbations. We further find the adversarial perturbations are either those tokens not existing in the training set or superficial cues associated with relation labels. To some extent, our approach unveils the characters against adversarial samples. We release an open-source testbed, ``\emph{DiagnoseAdv}''\footnote{The code and dataset are available in \url{https://github.com/zjunlp/DiagnoseAdv}.}, for future research purposes.
 
%Recent neural-based relation extraction approaches, though achieving promising improvement on benchmark datasets, have reported their vulnerability towards adversarial attacks. Thus far, efforts mostly focused on generating adversarial samples or defending adversarial attacks, but little is known about the difference between normal and adversarial samples. In this work, we take the first step to leverage the salience-based method to analyze those adversarial samples. We observe that salience tokens have a direct correlation with adversarial perturbations. We further find the adversarial perturbations are either those tokens not existing in the training set or superficial cues associated with relation labels. To some extent, our approach unveils the characters against adversarial samples. We release an open-source testbed, ``\emph{DiagnoseAdv}''\footnote{Work in Progress. The code and dataset are available in \url{https://anonymous.4open.science/r/25d76ae5-2dde-4c70-aa2d-cd0df40a9e66/}.}, for future research purposes.

\noindent\let\thefootnote\relax\footnotetext{
$*$ Equal contribution and shared co-first authorship. \\
$\star$ Corresponding author.
}
\end{abstract}
%%x




\iffalse
\begin{teaserfigure}
  \includegraphics[width=\textwidth]{sampleteaser}
  \caption{Seattle Mariners at Spring Training, 2010.}
  \Description{Enjoying the baseball game from the third-base
  seats. Ichiro Suzuki preparing to bat.}
  \label{fig:teaser}
\end{teaserfigure}
\fi
\maketitle

\section{Introduction} 
Relation Extraction (RE), aiming to extract the relation between two given entities based on their related context, is an important task for knowledge graph construction \cite{DBLP:conf/www/ZhangDSCZC20} which can benefit widespread domains such recommendation system \cite{DBLP:journals/jmpt/JiaZH20}, healthcare system \cite{DBLP:journals/corr/abs-2008-10813,DBLP:conf/emnlp/ZhangDLCZC20}, stock prediction \cite{DBLP:conf/www/DengZZCPC19} and so on. 
Previous neural-based models \cite{DBLP:conf/coling/ZengLLZZ14,zhang-etal-2018-attention,DBLP:conf/naacl/ZhangDSWCZC19,DBLP:conf/wsdm/DengZKZZC20,DBLP:conf/coling/LiWZZYC20,DBLP:journals/corr/abs-2009-07022,DBLP:conf/emnlp/ZhangDBYYCHZC20,DBLP:journals/corr/abs-2009-09841,DBLP:conf/coling/YuZDYZC20,DBLP:journals/corr/abs-2009-06207} have achieved promising performance on benchmark datasets, yet they are vulnerable to adversarial examples \cite{DBLP:conf/aaai/JinJZS20,DBLP:journals/tist/ZhangSAL20,DBLP:journals/corr/abs-2009-06206}. 

The study of adversarial examples and training ushered in a new era to understand and improve natural language processing (NLP) models.
However, recent approaches mainly focus on generating adversarial examples \cite{DBLP:conf/ndss/LiJDLW19,DBLP:conf/sp/GaoLSQ18,DBLP:conf/ijcai/0002LSBLS18} or defending adversarial attacks \cite{DBLP:conf/wsdm/EntezariADP20,DBLP:conf/cvpr/TheagarajanCBZ19}, the major difference between normal and adversarial samples is still not well-understood. 
Note that understanding adversarial examples can figure out missing connections of RE models and inspire important future studies \cite{DBLP:journals/tacl/BelinkovG19}. 
To this end, we formulate the following interesting research questions:  
\begin{quote}
    \emph{1. What is the difference between normal and adversarial samples?}\\
    \emph{2. What is the reason that adversarial examples mislead the prediction?}
\end{quote}
 
Motivated by this, we leverage integrated gradients \cite{DBLP:conf/icml/SundararajanTY17} to analyze the adversarial samples for RE. 
Firstly, we observe that salience tokens have a direct correlation with adversarial perturbations.
We then analyze the salience distribution of normal and adversarial samples and find that these salience distributions change slightly (\S~\ref{sec1}). 
Secondly, we conduct experiments to probe reasons for misclassification and find that the salience tokens of adversarial samples are either not existing in the training set or superficial cues associated with relation labels (\S~\ref{sec2}). 
In summary, our main contributions include:

\begin{itemize}
    \item To the best of our knowledge, we are the first to leverage salience-based analysis for adversarial samples in NLP, which provides a new perspective of understanding the model robustness. 
    \item We propose a simple yet effective method to probe adversarial samples with salience analysis and observe new findings that may promote future researches. 
    \item We provide an open-source testbed, ``\emph{DiagnoseAdv}'',  for future research purposes. Our framework can be readily applied to other NLP tasks such as text classification and sentiment analysis. 
\end{itemize}

\section{Analyzing Adversarial Samples for RE}
\subsection{Setup}
RE is usually formulated as a sequence classification problem. 
Formally, let $X=\left\{x_{1}, x_{2}, \ldots, x_{L}\right\}$ be an input sequence, $h,t \in X$ be two entities, and $Y$ be the output relations. 
The goal of this task is to estimate the conditional probability, $P(Y|X) = P(y|X,h,t)$

In this paper, we respectively leverage the pre-trained BERT \cite{bert} and MTB \cite{baldini-soares-etal-2019-matching} as the target model. 
Certainly, other strong models (e.g., SpanBERT \cite{DBLP:journals/tacl/JoshiCLWZL20} and XLNet \cite{DBLP:conf/nips/YangDYCSL19}) can also be leveraged. 
We preprocess the sentence, $\mathbf{x}=$ $\{w_1,$ $w_2,$ $h,$ $\dots,$ $t$, $\dots$,$w_L\}$, for the input form of BERT: $\mathbf{x}=$ $\{$[CLS]$,$ $w_1,$ $w_2,$ [E1], $h,$ [/E1], $\dots,$ [E2], $t,$ [/E2],..., $w_L,$ [SEP]$\}$, where $w_i, (i \in  [1, n])$ refers to each word in a sentence and $h$ as well as $t$ are head and tail entities, respectively. [E1], [/E1], [E2], and [/E2] are four special tokens used to mark the positions of the entities. 
Our approach can be readily applied to other classification tasks such as text classification and sentiment analysis. 

\subsection{Entity-aware Adversarial Attack}
We introduce an \textbf{entity-aware} adversarial attack method for RE in this section, where entities in original samples should not be changed during the adversarial attack.
Given a set of $N$ instances, $\mathcal{X}= \{X_1, X_2,\dots, X_N\}$ with a corresponding set of labels, $\mathcal{Y}=\{Y_1, Y_2,\dots, Y_N$\}, we have a RE model trained via the input $\mathcal{X}$ and $\mathcal{Y}$, which satisfies the formula $ \mathcal{Y}  = RE(\mathcal{X})$.

The adversarial example $X_{\mathrm{adv}}$ for each sentence $X \in \mathcal{X}$  should conform to the requirements as follows: 
\begin{equation}
\small
RE\left(X_{\mathrm{adv}} \right) \neq RE(X), \text { and } \operatorname{Sim}\left(X_{\mathrm{adv}}, X\right) \geq \epsilon,
\label{eq:ad_requirement}
\end{equation}
where $\mathrm{Sim}$ is a similarity function  and $\epsilon$ is the minimum similarity between the original and adversarial examples.
Note that $X_{\mathrm{adv}}$ should have the same entity pair as $\mathcal{X}$, thus, we constrain the entity token from being perturbed and extend both score-based adversarial attack approaches: TextFooler \cite{DBLP:conf/aaai/JinJZS20}, PWWS \cite{DBLP:conf/acl/RenDHC19}, and a gradient-based method: HotFlip \cite{DBLP:conf/acl/EbrahimiRLD18} in our experiment.
Other attack methods such as SememePSO \cite{zang-etal-2020-word}, TextBugger \cite{DBLP:conf/ndss/LiJDLW19}, UAT \cite{wallace-etal-2019-universal} can also be leveraged.

\subsection{Salience-based Analysis}
We leverage integrated gradients \cite{DBLP:conf/icml/SundararajanTY17} (IG) to analyze the identify inputs relevant to the prediction.
Attention-based attribution \cite{DBLP:conf/emnlp/WiegreffeP19} is not adopted as \citet{bastings-filippova-2020-elephant} point out saliency methods are more suitable than attention mechanism in providing faithful explanations. \citet{attention-not-commonsense} also notice that attention weights are insufficient when investigating the behavior of the attention head.
Among the saliency methods, the IG method is a variation from the gradient method that assigns importance by computing gradients of the output w.r.t. the input. IG outperforms simple gradient by dealing with the gradient \textit{saturation} problem that gradients may get close to zero when the function is well-fitted. Given an input sentence's embeddings $\mathbf{x}=\left\langle\mathbf{x}_{1}, \ldots, \mathbf{x}_{n}\right\rangle$ with $\mathbf{x}_i$ being embedding of the $i$-th input token, and a model $F$, we compute:
\begin{equation}
\small
\rm{IG}(\mathbf{x}_i)=\frac{1}{m} \sum_{j=1}^{m} \nabla_{\mathbf{x}_{i}} F\left(\mathbf{b}+\frac{j}{m}\left(\mathbf{x}-\mathbf{b}\right)\right) \cdot\left(\mathbf{x}_{i}-\mathbf{b}_{i}\right),
\end{equation}
where $\mathbf{b}$ is a baseline value, which is an all-zeros vector in our experiment. By averaging over gradients with linearly interpolated inputs between the baseline and the original input $\mathbf{x}$ in $m$ steps, and taking the dot product of the averaged gradient with the input embedding $\mathbf{x}_i$ minus the baseline, we get IG vectors for input tokens. In our experiment, we then use the norm of IG vectors as tokens' attribution scores.

\begin{figure*}[h] \centering
  \includegraphics[width=1\textwidth]{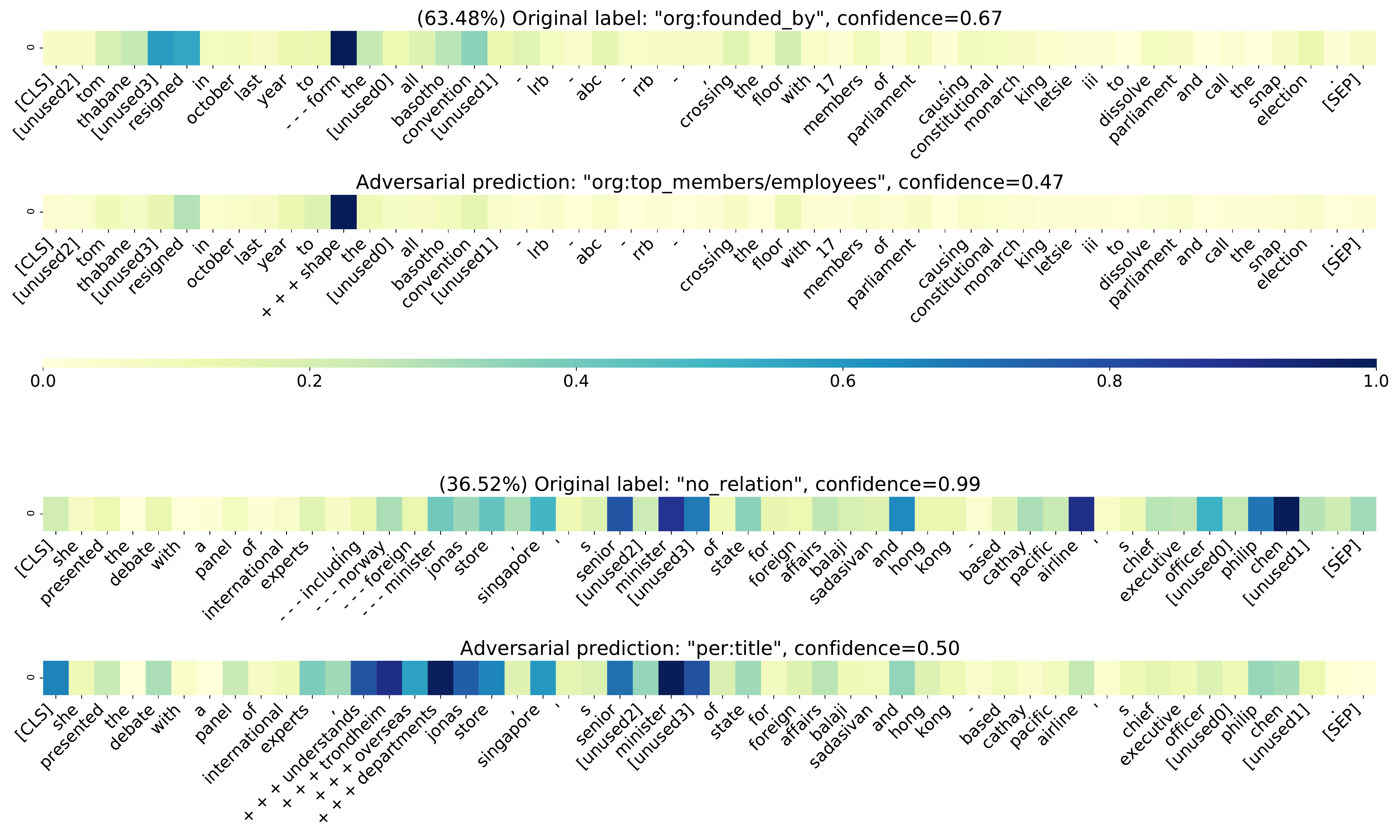}
\caption{Visualization of two types of how salience scores interact with perturbed tokens between normal samples and adversarial samples in TACRED.
The \emph{-\ -\ -} and \emph{+\ +\ +} signs mark perturbed tokens, representing token deletion in the original sample and insertion in the adversarial sample, respectively.}
\label{case1}
\end{figure*}
\section{Experiments}

We conduct experiments on two benchmark datasets: Wiki80\footnote{\url{https://github.com/thunlp/OpenNRE}} \cite{DBLP:conf/emnlp/HanZYWYLS18} and TACRED\footnote{\url{https://nlp.stanford.edu/projects/tacred/}} \cite{DBLP:conf/emnlp/ZhangZCAM17}. 
The Wiki80 dataset consisted of 80 relations, each having 700 instances. TACRED  is a large-scale RE dataset covering 42 relation types with 106,264 sentences. We provide an online GoogleColab for reproducibility\footnote{\url{https://colab.research.google.com/drive/1d4ayfzV8wqmGz0AxA1iLORfrD3JtbfYJ?usp=sharing}}.

% \subsection{Normal Samples vs. Adversarial Samples}
\subsection{What's Changed in Normal Samples?}
\label{sec1}
We conduct adversarial attacks to RE models as shown in Table \ref{adv_res}. We notice more adversarial samples are generated on the BERT model, indicating less vulnerability; among all three methods, HotFlip is most inefficient with success rates lower than 10\%.
To address \textbf{Question 1}, we leverage a token matching algorithm  to explore connections between the original and adversarial samples.

\begin{table}[h]
   %\fontsize{8}{10}\selectfont 
   \centering 
\resizebox{\linewidth}{!}{
\begin{tabular}{c|c|c}
\toprule
Model& \textbf{Wiki80} & \textbf{TACRED}\\
\midrule
BERT (Origin)&55,193/86.2 &99,008/67.5 \\
MTB (Origin)&55,225/90.3 &98,245/68.7 \\
 \midrule
BERT (HotFlip)&4,819/8.73\% &4,953/5.00\% \\
BERT (PWWS)&17,742/32.15\% &27,476/27.75\% \\
BERT (TextFooler)&26,774/48.51\% &34,892/35.24\% \\
 \midrule
MTB (HotFlip)&4,655/8.43\% &3,868/3.94\% \\
MTB (PWWS)&16,868/30.54\% &21,692/22.08\% \\
MTB (TextFooler)&25,969/47.02\% &25,751/26.21\% \\
 \bottomrule
\end{tabular}
}
\caption{Adversarial attack results from Wiki80 and TACRED dataset. The first two rows show numbers of correctly predicted samples and test performance (accuracy for Wiki80 and micro F1 for TACRED) of BERT or MTB model on two datasets, and the following rows indicate numbers of adversarial samples generated / success rate of adversarial attack with each (model, adversarial method) pair on each dataset.}
  \label{adv_res}
\end{table}
At sentence level, we have summarized two types of adversarial samples in Figure \ref{case1}: 1) the first type involves perturbations of $n$ tokens with highest salience scores in the original samples (except the irreplaceable entity tokens), while 2) the other type consists of samples in which no tokens with top salience scores are perturbed in these samples ($n=3$ in our experiment).
The ratio of samples in the first type greatly exceeds the second one among different adversarial methods on each dataset.
\begin{figure}[h] \centering
  \includegraphics[width=0.50\textwidth]{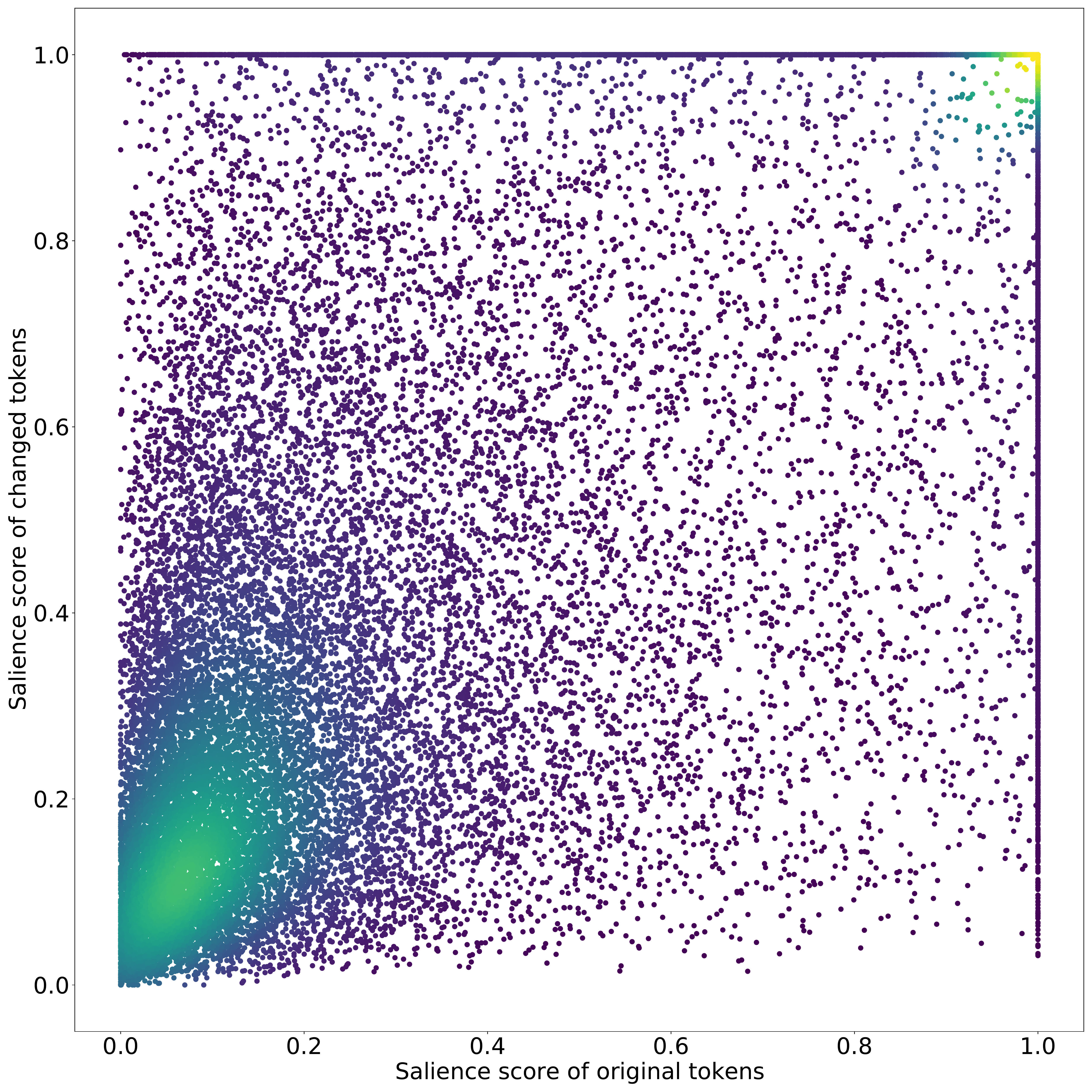}
\caption{Salience score changes of perturbed positions during the TextFooler attack in Wiki80. The X and Y-axis coordinates stand for salience scores of perturbed positions in original samples and adversarial samples.}
\label{normal_adv}
\end{figure}

\begin{figure}[h] \centering
  \includegraphics[width=0.50\textwidth]{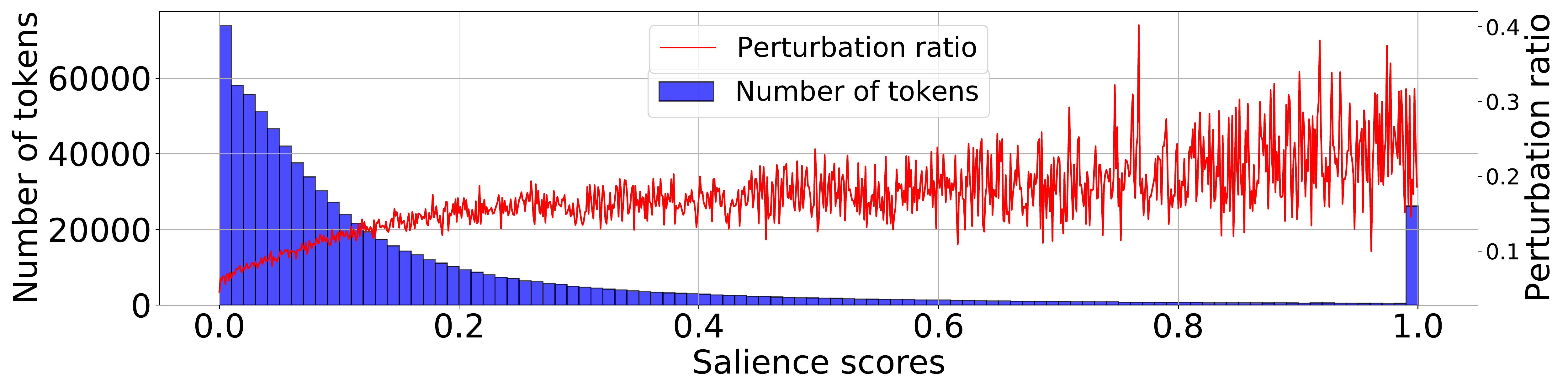}
\caption{Tokens' distribution and perturbation ratio along salience scores of Wiki80.}
\label{token_dist}
\end{figure}
At a finer-grained token level, we explore salience scores of tokens at perturbed positions as shown in Figure \ref{normal_adv}. 
Each point represents a perturbed position, whose X-axis and Y-axis coordinate stand for its salience score in the original sample and the adversarial sample, respectively. Most points scatter along the diagonal $y=x$, indicating the stability of tokens' influence on predictions before and after being perturbed.
Colors of points indicate one largest cluster around (0.05, 0.05) and the second-largest cluster around (1, 1).
This phenomenon can be explained by Figure \ref{token_dist}, which reveals the distribution of all tokens in the original samples whose salience scores are mostly around 0.05 and 1.0.
It also reveals that although above 2/3 samples in the original sample involve perturbations of tokens with the top salience scores s, most perturbed tokens have low salience scores in token-level. However, from the perturbation ratio curve in Figure \ref{token_dist},  tokens with higher salience scores are more likely to be perturbed.

In conclusion, we observe the strong correlation between perturbations in the adversarial samples and high salience scores in the original samples, which is intuitive as high salience scores reflect tokens' impact on the model's predictions, perturbing those tokens are likely to change the predictions. We also argue that current adversarial methods are inefficient in RE, as they perturb many low-salience tokens in the original samples.

\subsection{Why MisClassified?}
% \subsection{What's Changed in Adversarial Samples?}
\label{sec2}
To address \textbf{Question 2} and further analyze why the model predicts differently with few perturbations, we look into the perturbed tokens in the adversarial samples.
\iffalse
\begin{figure}[H] \centering
  \includegraphics[width=0.45\textwidth]{figs/confidence.png}
\caption{The distribution of confidence drop.}
\label{confidence}
\end{figure}
\fi
\begin{table}[t]
    \centering
    \small
    \begin{tabular}{p{0.93\linewidth}}
    \toprule
          %\multicolumn{1}{c}{\textbf{C+M}}\\
    %\midrule
     Actress \textbf{Mia Farrow}  \textit{\color{red}had} \textbf{Vidal Sassoon} give her the look when she married Frank Sinatra in 1966, and she also wore it in her 1968 film ``rosemary's baby.''\\
    
         \textbf{\color{ourdarkgreen}Label}: \textbf{\texttt{no\_relation}}\\
        \textbf{\color{ourdarkblue}Prediction}: \textbf{\texttt{no\_relation}}\\
        
        \specialrule{0em}{4pt}{4pt}

        Actress \textbf{Mia Farrow}  \textit{\color{red}birth} \textbf{Vidal Sassoon} give her the look when she married Frank Sinatra in 1966, and she also wore it in her 1968 film ``rosemary's baby.''\\
        
         \textbf{\color{ourdarkgreen}Label}: \textbf{\texttt{no\_relation}}\\
        \textbf{\color{ourdarkblue}Prediction}: \textbf{\texttt{per:parents}}\\
        
    \bottomrule
    \end{tabular}
    \caption{Predictions on normal (above) and adversarial samples (bellow), where \textbf{bold} tokens are entities and {\color{red}red} represents perturbed tokens. We can observe that those perturbed tokens have superficial cues associated with corresponding relation labels.}
    \label{case2}
   % \vspace{-1em}
\end{table}

We manually examine perturbed tokens with high salience scores in the adversarial samples and observe a high ratio of superficial association between the predictions and the perturbed tokens, i.e., the model makes a wrong prediction upon seeing a frequent co-word. For example, as shown in Table \ref{case2}, the perturbed token \emph{birth} has a spurious correlation with the predicted label \emph{per:parents} in train samples, thus leading to the misclassification. We have examined 3,868 adversarial samples in TACRED (MTB, HotFlip). Such association accounts for 2,248 (58.12\%) adversarial samples, reflecting that neural networks tend to capture co-occurrence information between the token and label while ignoring low-frequency but important causal information. We argue that such artifacts and spurious correlation in the data mainly mislead the classification of the adversarial samples \cite{han-etal-2020-explaining}.

We also notice around 40\% adversarial samples contain perturbed tokens that do not appear in the training set, which leads to the input being Out-Of-Distribution (OOD). 
% Although it is much alleviated by word-piece tokenization, 
We also abserve that the OOD problem results are accompanied by a decrease in confidence, revealing that OOD problem may be annother minor reason for misclassification.

\subsection{Extra Statistics of Adversarial Samples}
%In this section, we present extra statistics of generated adversarial samples as follows:
\begin{table}[h]
%   \fontsize{8}{10}\selectfont 
   \centering 
\resizebox{\linewidth}{!}{
\begin{tabular}{c|c|c|c|c}
\toprule
Model& \textbf{Avg. Perturb} &\textbf{\% Salience} &\textbf{\% OOD} &\textbf{Avg. Confidence} \\
\midrule
BERT (HotFlip) & 6.72 & 91.99 & 49.47 & -0.24\\
BERT (PWWS) & 4.42 & 91.15 & 41.05 & -0.25\\
BERT (TextFooler) & 3.72 & 83.66 & 42.28 & -0.29\\
\midrule
MTB (HotFlip) & 6.65 & 91.69 & 48.46 & -0.28\\
MTB (PWWS) & 4.31 & 90.66 & 39.63 & -0.30\\
MTB (TextFooler) & 3.66 & 83.39 & 40.92 & -0.33\\
 \bottomrule
\end{tabular}
}
\caption{Extra statistics of Wiki80 adversarial samples.}
  \label{adv_res_wiki}
\end{table}

\begin{table}[h]
%   \fontsize{8}{10}\selectfont 
   \centering 
\resizebox{\linewidth}{!}{
\begin{tabular}{c|c|c|c|c}
\toprule
Model& \textbf{Avg. Perturb} &\textbf{\% Salience} &\textbf{\% OOD} &\textbf{Avg. Confidence} \\
\midrule
BERT (HotFlip) & 6.70 & 64.75 & 50.80 & -0.17\\
BERT (PWWS) & 4.70 & 74.20 & 40.50 & -0.27\\
BERT (TextFooler) & 4.69 & 63.48 & 54.88 & -0.36\\
\midrule
MTB (HotFlip) & 6.86 & 68.95 & 51.16 & -0.16\\
MTB (PWWS) & 4.72 & 79.17 & 41.50 & -0.25\\
MTB (TextFooler) & 4.67 & 71.87 & 53.82 & -0.32\\
 \bottomrule
\end{tabular}
}
\caption{Extra statistics of TACRED adversarial samples.}
  \label{adv_res_tacred}
\end{table}

In the Table \ref{adv_res_wiki} and \ref{adv_res_tacred}, the column ``Avg. Perturb" refers to average token perturbations from original samples, ``\% Salience" refers to the ratio of adversarial samples involving perturbations of relatively high salience scores (top 3 highest except the entity tokens), ``\% OOD" means ratio of samples containing Out-Of-Distribution tokens, and``Avg. Confidence" refers to the average decrease of prediction confidence between adversarial samples and of original samples (minus values mean lower confidence in adversarial samples).

% We filter those adversarial examples with high confidence (confidence $\ge$ 0.8).
% Then, we analyze the correlation of salience tokens and relation labels.
% We find that those salience tokens have a high correlation with those labels. 
% We find that nearly 80\% of the high confidence adversarial samples have this phenomenon. 
% Note that neural networks tend to capture co-occurrence information between the token and label while ignoring low-frequency but important causal information. 
% We argue that such artifacts in the data mislead the classification \cite{han-etal-2020-explaining}. 

% We further analyze those adversarial examples with low confidence. We observe that 70\% of perturbed tokens in adversarial samples do not exist in the train set. We think those tokens may lead to the input being Out-of-Distribution (OOD), which lowers the confidence to change the model's predictions. 

% In conclusion, we believe spurious correlation is one main reason for misclassification of the adversarial samples, and we regard the OOD problem as a weaker factor for decrease in model's prediction confidence on adversarial samples.

\section{Conclusion}
We introduce the entity-aware adversarial attack for Relation Extraction, and leverage the salience-based analysis of adversarial samples. We observe that correlation between high salience scores with token perturbations, inspiring future works of salience-aware data augmentation. 
Furthermore, we identify two factors: spurious correlation and OOD as main reasons for adversarial misclassification.
Breaking down the spurious correlation with causal analysis may help defend adversarial attacks with better generalization. 
More future works should also be taken into consideration for those OOD samples. 
We regard this study as a small step towards the understanding of adversarial samples. 

\begin{acks}
This work is funded by NSFC91846204/NSFCU19B2027.
\end{acks}
%\section*{Broader Impact Statement}
%Neural networks have achieved great success in a wide range of NLP applications, such as machine translation, question answering, dialogue systems, etc. Despite their success, the wide adoption of neural networks in real-world missions is hindered by the security concerns of neural networks because slight, imperceptible perturbations are capable of causing incorrect behaviors of neural networks. Our work focuses on unveiling adversarial samples' characters, promoting developing more robust models, and benefit lots of real-world applications. 

\bibliographystyle{ACM-Reference-Format}
\bibliography{sample-base}


\begin{thebibliography}{39}

%%% the Web address.

\ifx \showCODEN    \undefined \def \showCODEN     #1{\unskip}     \fi
\ifx \showDOI      \undefined \def \showDOI       #1{#1}\fi
\ifx \showISBNx    \undefined \def \showISBNx     #1{\unskip}     \fi
\ifx \showISBNxiii \undefined \def \showISBNxiii  #1{\unskip}     \fi
\ifx \showISSN     \undefined \def \showISSN      #1{\unskip}     \fi
\ifx \showLCCN     \undefined \def \showLCCN      #1{\unskip}     \fi
\ifx \shownote     \undefined \def \shownote      #1{#1}          \fi
\ifx \showarticletitle \undefined \def \showarticletitle #1{#1}   \fi
\ifx \showURL      \undefined \def \showURL       {\relax}        \fi
% invisible to TeX
\providecommand\bibfield[2]{#2}
\providecommand\bibinfo[2]{#2}
\providecommand\natexlab[1]{#1}
\providecommand\showeprint[2][]{arXiv:#2}

\bibitem[\protect\citeauthoryear{Baldini~Soares, FitzGerald, Ling, and
  Kwiatkowski}{Baldini~Soares et~al\mbox{.}}{2019}]%
        {baldini-soares-etal-2019-matching}
\bibfield{author}{\bibinfo{person}{Livio Baldini~Soares},
  \bibinfo{person}{Nicholas FitzGerald}, \bibinfo{person}{Jeffrey Ling}, {and}
  \bibinfo{person}{Tom Kwiatkowski}.} \bibinfo{year}{2019}\natexlab{}.
\newblock \showarticletitle{Matching the Blanks: Distributional Similarity for
  Relation Learning}. In \bibinfo{booktitle}{\emph{Proceedings of the 57th
  Annual Meeting of the Association for Computational Linguistics}}.
  \bibinfo{publisher}{Association for Computational Linguistics},
  \bibinfo{address}{Florence, Italy}, \bibinfo{pages}{2895--2905}.
\newblock
\urldef\tempurl%
\url{https://doi.org/10.18653/v1/P19-1279}
\showDOI{\tempurl}


\bibitem[\protect\citeauthoryear{Bastings and Filippova}{Bastings and
  Filippova}{2020}]%
        {bastings-filippova-2020-elephant}
\bibfield{author}{\bibinfo{person}{Jasmijn Bastings} {and}
  \bibinfo{person}{Katja Filippova}.} \bibinfo{year}{2020}\natexlab{}.
\newblock \showarticletitle{The elephant in the interpretability room: Why use
  attention as explanation when we have saliency methods?}. In
  \bibinfo{booktitle}{\emph{Proceedings of the Third BlackboxNLP Workshop on
  Analyzing and Interpreting Neural Networks for NLP}}.
  \bibinfo{publisher}{Association for Computational Linguistics},
  \bibinfo{address}{Online}, \bibinfo{pages}{149--155}.
\newblock
\urldef\tempurl%
\url{https://doi.org/10.18653/v1/2020.blackboxnlp-1.14}
\showDOI{\tempurl}


\bibitem[\protect\citeauthoryear{Belinkov and Glass}{Belinkov and
  Glass}{2019}]%
        {DBLP:journals/tacl/BelinkovG19}
\bibfield{author}{\bibinfo{person}{Yonatan Belinkov} {and}
  \bibinfo{person}{James~R. Glass}.} \bibinfo{year}{2019}\natexlab{}.
\newblock \showarticletitle{Analysis Methods in Neural Language Processing: {A}
  Survey}.
\newblock \bibinfo{journal}{\emph{Trans. Assoc. Comput. Linguistics}}
  \bibinfo{volume}{7} (\bibinfo{year}{2019}), \bibinfo{pages}{49--72}.
\newblock
\urldef\tempurl%
\url{https://transacl.org/ojs/index.php/tacl/article/view/1570}
\showURL{%
\tempurl}


\bibitem[\protect\citeauthoryear{Deng, Zhang, Kang, Zhang, Zhang, and
  Chen}{Deng et~al\mbox{.}}{2020}]%
        {DBLP:conf/wsdm/DengZKZZC20}
\bibfield{author}{\bibinfo{person}{Shumin Deng}, \bibinfo{person}{Ningyu
  Zhang}, \bibinfo{person}{Jiaojian Kang}, \bibinfo{person}{Yichi Zhang},
  \bibinfo{person}{Wei Zhang}, {and} \bibinfo{person}{Huajun Chen}.}
  \bibinfo{year}{2020}\natexlab{}.
\newblock \showarticletitle{Meta-Learning with Dynamic-Memory-Based
  Prototypical Network for Few-Shot Event Detection}. In
  \bibinfo{booktitle}{\emph{{WSDM} '20: The Thirteenth {ACM} International
  Conference on Web Search and Data Mining, Houston, TX, USA, February 3-7,
  2020}}, \bibfield{editor}{\bibinfo{person}{James Caverlee},
  \bibinfo{person}{Xia~(Ben) Hu}, \bibinfo{person}{Mounia Lalmas}, {and}
  \bibinfo{person}{Wei Wang}} (Eds.). \bibinfo{publisher}{{ACM}},
  \bibinfo{pages}{151--159}.
\newblock
\urldef\tempurl%
\url{https://doi.org/10.1145/3336191.3371796}
\showDOI{\tempurl}


\bibitem[\protect\citeauthoryear{Deng, Zhang, Zhang, Chen, Pan, and Chen}{Deng
  et~al\mbox{.}}{2019}]%
        {DBLP:conf/www/DengZZCPC19}
\bibfield{author}{\bibinfo{person}{Shumin Deng}, \bibinfo{person}{Ningyu
  Zhang}, \bibinfo{person}{Wen Zhang}, \bibinfo{person}{Jiaoyan Chen},
  \bibinfo{person}{Jeff~Z. Pan}, {and} \bibinfo{person}{Huajun Chen}.}
  \bibinfo{year}{2019}\natexlab{}.
\newblock \showarticletitle{Knowledge-Driven Stock Trend Prediction and
  Explanation via Temporal Convolutional Network}. In
  \bibinfo{booktitle}{\emph{Companion of The 2019 World Wide Web Conference,
  {WWW} 2019, San Francisco, CA, USA, May 13-17, 2019}},
  \bibfield{editor}{\bibinfo{person}{Sihem Amer{-}Yahia},
  \bibinfo{person}{Mohammad Mahdian}, \bibinfo{person}{Ashish Goel},
  \bibinfo{person}{Geert{-}Jan Houben}, \bibinfo{person}{Kristina Lerman},
  \bibinfo{person}{Julian~J. McAuley}, \bibinfo{person}{Ricardo Baeza{-}Yates},
  {and} \bibinfo{person}{Leila Zia}} (Eds.). \bibinfo{publisher}{{ACM}},
  \bibinfo{pages}{678--685}.
\newblock
\urldef\tempurl%
\url{https://doi.org/10.1145/3308560.3317701}
\showDOI{\tempurl}


\bibitem[\protect\citeauthoryear{Devlin, Chang, Lee, and Toutanova}{Devlin
  et~al\mbox{.}}{2019}]%
        {bert}
\bibfield{author}{\bibinfo{person}{Jacob Devlin}, \bibinfo{person}{Ming-Wei
  Chang}, \bibinfo{person}{Kenton Lee}, {and} \bibinfo{person}{Kristina
  Toutanova}.} \bibinfo{year}{2019}\natexlab{}.
\newblock \showarticletitle{{BERT}: Pre-training of Deep Bidirectional
  Transformers for Language Understanding}. In
  \bibinfo{booktitle}{\emph{Proceedings of the 2019 Conference of the North
  {A}merican Chapter of the Association for Computational Linguistics: Human
  Language Technologies, Volume 1 (Long and Short Papers)}}.
  \bibinfo{publisher}{Association for Computational Linguistics},
  \bibinfo{address}{Minneapolis, Minnesota}, \bibinfo{pages}{4171--4186}.
\newblock
\urldef\tempurl%
\url{https://doi.org/10.18653/v1/N19-1423}
\showDOI{\tempurl}


\bibitem[\protect\citeauthoryear{Ebrahimi, Rao, Lowd, and Dou}{Ebrahimi
  et~al\mbox{.}}{2018}]%
        {DBLP:conf/acl/EbrahimiRLD18}
\bibfield{author}{\bibinfo{person}{Javid Ebrahimi}, \bibinfo{person}{Anyi Rao},
  \bibinfo{person}{Daniel Lowd}, {and} \bibinfo{person}{Dejing Dou}.}
  \bibinfo{year}{2018}\natexlab{}.
\newblock \showarticletitle{HotFlip: White-Box Adversarial Examples for Text
  Classification}. In \bibinfo{booktitle}{\emph{Proceedings of the 56th Annual
  Meeting of the Association for Computational Linguistics, {ACL} 2018,
  Melbourne, Australia, July 15-20, 2018, Volume 2: Short Papers}},
  \bibfield{editor}{\bibinfo{person}{Iryna Gurevych} {and}
  \bibinfo{person}{Yusuke Miyao}} (Eds.). \bibinfo{publisher}{Association for
  Computational Linguistics}, \bibinfo{pages}{31--36}.
\newblock
\urldef\tempurl%
\url{https://doi.org/10.18653/v1/P18-2006}
\showDOI{\tempurl}


\bibitem[\protect\citeauthoryear{Entezari, Al{-}Sayouri, Darvishzadeh, and
  Papalexakis}{Entezari et~al\mbox{.}}{2020}]%
        {DBLP:conf/wsdm/EntezariADP20}
\bibfield{author}{\bibinfo{person}{Negin Entezari}, \bibinfo{person}{Saba~A.
  Al{-}Sayouri}, \bibinfo{person}{Amirali Darvishzadeh}, {and}
  \bibinfo{person}{Evangelos~E. Papalexakis}.} \bibinfo{year}{2020}\natexlab{}.
\newblock \showarticletitle{All You Need Is Low (Rank): Defending Against
  Adversarial Attacks on Graphs}. In \bibinfo{booktitle}{\emph{{WSDM} '20: The
  Thirteenth {ACM} International Conference on Web Search and Data Mining,
  Houston, TX, USA, February 3-7, 2020}},
  \bibfield{editor}{\bibinfo{person}{James Caverlee},
  \bibinfo{person}{Xia~(Ben) Hu}, \bibinfo{person}{Mounia Lalmas}, {and}
  \bibinfo{person}{Wei Wang}} (Eds.). \bibinfo{publisher}{{ACM}},
  \bibinfo{pages}{169--177}.
\newblock
\urldef\tempurl%
\url{https://doi.org/10.1145/3336191.3371789}
\showDOI{\tempurl}


\bibitem[\protect\citeauthoryear{Gao, Lanchantin, Soffa, and Qi}{Gao
  et~al\mbox{.}}{2018}]%
        {DBLP:conf/sp/GaoLSQ18}
\bibfield{author}{\bibinfo{person}{Ji Gao}, \bibinfo{person}{Jack Lanchantin},
  \bibinfo{person}{Mary~Lou Soffa}, {and} \bibinfo{person}{Yanjun Qi}.}
  \bibinfo{year}{2018}\natexlab{}.
\newblock \showarticletitle{Black-Box Generation of Adversarial Text Sequences
  to Evade Deep Learning Classifiers}. In \bibinfo{booktitle}{\emph{2018 {IEEE}
  Security and Privacy Workshops, {SP} Workshops 2018, San Francisco, CA, USA,
  May 24, 2018}}. \bibinfo{publisher}{{IEEE} Computer Society},
  \bibinfo{pages}{50--56}.
\newblock
\urldef\tempurl%
\url{https://doi.org/10.1109/SPW.2018.00016}
\showDOI{\tempurl}


\bibitem[\protect\citeauthoryear{Han, Wallace, and Tsvetkov}{Han
  et~al\mbox{.}}{2020}]%
        {han-etal-2020-explaining}
\bibfield{author}{\bibinfo{person}{Xiaochuang Han}, \bibinfo{person}{Byron~C.
  Wallace}, {and} \bibinfo{person}{Yulia Tsvetkov}.}
  \bibinfo{year}{2020}\natexlab{}.
\newblock \showarticletitle{Explaining Black Box Predictions and Unveiling Data
  Artifacts through Influence Functions}. In
  \bibinfo{booktitle}{\emph{Proceedings of the 58th Annual Meeting of the
  Association for Computational Linguistics}}. \bibinfo{publisher}{Association
  for Computational Linguistics}, \bibinfo{address}{Online},
  \bibinfo{pages}{5553--5563}.
\newblock
\urldef\tempurl%
\url{https://doi.org/10.18653/v1/2020.acl-main.492}
\showDOI{\tempurl}


\bibitem[\protect\citeauthoryear{Han, Zhu, Yu, Wang, Yao, Liu, and Sun}{Han
  et~al\mbox{.}}{2018}]%
        {DBLP:conf/emnlp/HanZYWYLS18}
\bibfield{author}{\bibinfo{person}{Xu Han}, \bibinfo{person}{Hao Zhu},
  \bibinfo{person}{Pengfei Yu}, \bibinfo{person}{Ziyun Wang},
  \bibinfo{person}{Yuan Yao}, \bibinfo{person}{Zhiyuan Liu}, {and}
  \bibinfo{person}{Maosong Sun}.} \bibinfo{year}{2018}\natexlab{}.
\newblock \showarticletitle{FewRel: {A} Large-Scale Supervised Few-shot
  Relation Classification Dataset with State-of-the-Art Evaluation}. In
  \bibinfo{booktitle}{\emph{Proceedings of the 2018 Conference on Empirical
  Methods in Natural Language Processing, Brussels, Belgium, October 31 -
  November 4, 2018}}, \bibfield{editor}{\bibinfo{person}{Ellen Riloff},
  \bibinfo{person}{David Chiang}, \bibinfo{person}{Julia Hockenmaier}, {and}
  \bibinfo{person}{Jun'ichi Tsujii}} (Eds.). \bibinfo{publisher}{Association
  for Computational Linguistics}, \bibinfo{pages}{4803--4809}.
\newblock
\urldef\tempurl%
\url{https://doi.org/10.18653/v1/d18-1514}
\showDOI{\tempurl}


\bibitem[\protect\citeauthoryear{Jia, Zhang, and Hua}{Jia
  et~al\mbox{.}}{2020}]%
        {DBLP:journals/jmpt/JiaZH20}
\bibfield{author}{\bibinfo{person}{Qianghuai Jia}, \bibinfo{person}{Ningyu
  Zhang}, {and} \bibinfo{person}{Nengwei Hua}.}
  \bibinfo{year}{2020}\natexlab{}.
\newblock \showarticletitle{Context-Aware Deep Model for Entity Recommendation
  System in Search Engine at Alibaba}.
\newblock \bibinfo{journal}{\emph{J. Multim. Process. Technol.}}
  \bibinfo{volume}{11}, \bibinfo{number}{1} (\bibinfo{year}{2020}),
  \bibinfo{pages}{23--35}.
\newblock
\urldef\tempurl%
\url{https://doi.org/10.6025/jmpt/2020/11/1/23-35}
\showDOI{\tempurl}


\bibitem[\protect\citeauthoryear{Jin, Jin, Zhou, and Szolovits}{Jin
  et~al\mbox{.}}{2020}]%
        {DBLP:conf/aaai/JinJZS20}
\bibfield{author}{\bibinfo{person}{Di Jin}, \bibinfo{person}{Zhijing Jin},
  \bibinfo{person}{Joey~Tianyi Zhou}, {and} \bibinfo{person}{Peter Szolovits}.}
  \bibinfo{year}{2020}\natexlab{}.
\newblock \showarticletitle{Is {BERT} Really Robust? {A} Strong Baseline for
  Natural Language Attack on Text Classification and Entailment}. In
  \bibinfo{booktitle}{\emph{The Thirty-Fourth {AAAI} Conference on Artificial
  Intelligence, {AAAI} 2020, The Thirty-Second Innovative Applications of
  Artificial Intelligence Conference, {IAAI} 2020, The Tenth {AAAI} Symposium
  on Educational Advances in Artificial Intelligence, {EAAI} 2020, New York,
  NY, USA, February 7-12, 2020}}. \bibinfo{publisher}{{AAAI} Press},
  \bibinfo{pages}{8018--8025}.
\newblock
\urldef\tempurl%
\url{https://aaai.org/ojs/index.php/AAAI/article/view/6311}
\showURL{%
\tempurl}


\bibitem[\protect\citeauthoryear{Joshi, Chen, Liu, Weld, Zettlemoyer, and
  Levy}{Joshi et~al\mbox{.}}{2020}]%
        {DBLP:journals/tacl/JoshiCLWZL20}
\bibfield{author}{\bibinfo{person}{Mandar Joshi}, \bibinfo{person}{Danqi Chen},
  \bibinfo{person}{Yinhan Liu}, \bibinfo{person}{Daniel~S. Weld},
  \bibinfo{person}{Luke Zettlemoyer}, {and} \bibinfo{person}{Omer Levy}.}
  \bibinfo{year}{2020}\natexlab{}.
\newblock \showarticletitle{SpanBERT: Improving Pre-training by Representing
  and Predicting Spans}.
\newblock \bibinfo{journal}{\emph{Trans. Assoc. Comput. Linguistics}}
  \bibinfo{volume}{8} (\bibinfo{year}{2020}), \bibinfo{pages}{64--77}.
\newblock
\urldef\tempurl%
\url{https://transacl.org/ojs/index.php/tacl/article/view/1853}
\showURL{%
\tempurl}


\bibitem[\protect\citeauthoryear{Klein and Nabi}{Klein and Nabi}{2019}]%
        {attention-not-commonsense}
\bibfield{author}{\bibinfo{person}{Tassilo Klein} {and} \bibinfo{person}{Moin
  Nabi}.} \bibinfo{year}{2019}\natexlab{}.
\newblock \showarticletitle{Attention Is (not) All You Need for Commonsense
  Reasoning}. In \bibinfo{booktitle}{\emph{Proceedings of the 57th Annual
  Meeting of the Association for Computational Linguistics}}.
  \bibinfo{publisher}{Association for Computational Linguistics},
  \bibinfo{address}{Florence, Italy}, \bibinfo{pages}{4831--4836}.
\newblock
\urldef\tempurl%
\url{https://doi.org/10.18653/v1/P19-1477}
\showDOI{\tempurl}


\bibitem[\protect\citeauthoryear{Li, Ji, Du, Li, and Wang}{Li
  et~al\mbox{.}}{2019}]%
        {DBLP:conf/ndss/LiJDLW19}
\bibfield{author}{\bibinfo{person}{Jinfeng Li}, \bibinfo{person}{Shouling Ji},
  \bibinfo{person}{Tianyu Du}, \bibinfo{person}{Bo Li}, {and}
  \bibinfo{person}{Ting Wang}.} \bibinfo{year}{2019}\natexlab{}.
\newblock \showarticletitle{TextBugger: Generating Adversarial Text Against
  Real-world Applications}. In \bibinfo{booktitle}{\emph{26th Annual Network
  and Distributed System Security Symposium, {NDSS} 2019, San Diego,
  California, USA, February 24-27, 2019}}. \bibinfo{publisher}{The Internet
  Society}.
\newblock
\urldef\tempurl%
\url{https://www.ndss-symposium.org/ndss-paper/textbugger-generating-adversarial-text-against-real-world-applications/}
\showURL{%
\tempurl}


\bibitem[\protect\citeauthoryear{Li, Wang, Zhang, Zhang, Yang, and Chen}{Li
  et~al\mbox{.}}{2020}]%
        {DBLP:conf/coling/LiWZZYC20}
\bibfield{author}{\bibinfo{person}{Juan Li}, \bibinfo{person}{Ruoxu Wang},
  \bibinfo{person}{Ningyu Zhang}, \bibinfo{person}{Wen Zhang},
  \bibinfo{person}{Fan Yang}, {and} \bibinfo{person}{Huajun Chen}.}
  \bibinfo{year}{2020}\natexlab{}.
\newblock \showarticletitle{Logic-guided Semantic Representation Learning for
  Zero-Shot Relation Classification}. In \bibinfo{booktitle}{\emph{Proceedings
  of the 28th International Conference on Computational Linguistics, {COLING}
  2020, Barcelona, Spain (Online), December 8-13, 2020}},
  \bibfield{editor}{\bibinfo{person}{Donia Scott}, \bibinfo{person}{N{\'{u}}ria
  Bel}, {and} \bibinfo{person}{Chengqing Zong}} (Eds.).
  \bibinfo{publisher}{International Committee on Computational Linguistics},
  \bibinfo{pages}{2967--2978}.
\newblock
\urldef\tempurl%
\url{https://doi.org/10.18653/v1/2020.coling-main.265}
\showDOI{\tempurl}


\bibitem[\protect\citeauthoryear{Liang, Li, Su, Bian, Li, and Shi}{Liang
  et~al\mbox{.}}{2018}]%
        {DBLP:conf/ijcai/0002LSBLS18}
\bibfield{author}{\bibinfo{person}{Bin Liang}, \bibinfo{person}{Hongcheng Li},
  \bibinfo{person}{Miaoqiang Su}, \bibinfo{person}{Pan Bian},
  \bibinfo{person}{Xirong Li}, {and} \bibinfo{person}{Wenchang Shi}.}
  \bibinfo{year}{2018}\natexlab{}.
\newblock \showarticletitle{Deep Text Classification Can be Fooled}. In
  \bibinfo{booktitle}{\emph{Proceedings of the Twenty-Seventh International
  Joint Conference on Artificial Intelligence, {IJCAI} 2018, July 13-19, 2018,
  Stockholm, Sweden}}, \bibfield{editor}{\bibinfo{person}{J{\'{e}}r{\^{o}}me
  Lang}} (Ed.). \bibinfo{publisher}{ijcai.org}, \bibinfo{pages}{4208--4215}.
\newblock
\urldef\tempurl%
\url{https://doi.org/10.24963/ijcai.2018/585}
\showDOI{\tempurl}


\bibitem[\protect\citeauthoryear{Ren, Deng, He, and Che}{Ren
  et~al\mbox{.}}{2019}]%
        {DBLP:conf/acl/RenDHC19}
\bibfield{author}{\bibinfo{person}{Shuhuai Ren}, \bibinfo{person}{Yihe Deng},
  \bibinfo{person}{Kun He}, {and} \bibinfo{person}{Wanxiang Che}.}
  \bibinfo{year}{2019}\natexlab{}.
\newblock \showarticletitle{Generating Natural Language Adversarial Examples
  through Probability Weighted Word Saliency}. In
  \bibinfo{booktitle}{\emph{Proceedings of the 57th Conference of the
  Association for Computational Linguistics, {ACL} 2019, Florence, Italy, July
  28- August 2, 2019, Volume 1: Long Papers}},
  \bibfield{editor}{\bibinfo{person}{Anna Korhonen}, \bibinfo{person}{David~R.
  Traum}, {and} \bibinfo{person}{Llu{\'{\i}}s M{\`{a}}rquez}} (Eds.).
  \bibinfo{publisher}{Association for Computational Linguistics},
  \bibinfo{pages}{1085--1097}.
\newblock
\urldef\tempurl%
\url{https://doi.org/10.18653/v1/p19-1103}
\showDOI{\tempurl}


\bibitem[\protect\citeauthoryear{Sundararajan, Taly, and Yan}{Sundararajan
  et~al\mbox{.}}{2017}]%
        {DBLP:conf/icml/SundararajanTY17}
\bibfield{author}{\bibinfo{person}{Mukund Sundararajan}, \bibinfo{person}{Ankur
  Taly}, {and} \bibinfo{person}{Qiqi Yan}.} \bibinfo{year}{2017}\natexlab{}.
\newblock \showarticletitle{Axiomatic Attribution for Deep Networks}. In
  \bibinfo{booktitle}{\emph{Proceedings of the 34th International Conference on
  Machine Learning, {ICML} 2017, Sydney, NSW, Australia, 6-11 August 2017}}
  \emph{(\bibinfo{series}{Proceedings of Machine Learning Research})},
  \bibfield{editor}{\bibinfo{person}{Doina Precup} {and}
  \bibinfo{person}{Yee~Whye Teh}} (Eds.), Vol.~\bibinfo{volume}{70}.
  \bibinfo{publisher}{{PMLR}}, \bibinfo{pages}{3319--3328}.
\newblock
\urldef\tempurl%
\url{http://proceedings.mlr.press/v70/sundararajan17a.html}
\showURL{%
\tempurl}


\bibitem[\protect\citeauthoryear{Theagarajan, Chen, Bhanu, and
  Zhang}{Theagarajan et~al\mbox{.}}{2019}]%
        {DBLP:conf/cvpr/TheagarajanCBZ19}
\bibfield{author}{\bibinfo{person}{Rajkumar Theagarajan}, \bibinfo{person}{Ming
  Chen}, \bibinfo{person}{Bir Bhanu}, {and} \bibinfo{person}{Jing Zhang}.}
  \bibinfo{year}{2019}\natexlab{}.
\newblock \showarticletitle{ShieldNets: Defending Against Adversarial Attacks
  Using Probabilistic Adversarial Robustness}. In
  \bibinfo{booktitle}{\emph{{IEEE} Conference on Computer Vision and Pattern
  Recognition, {CVPR} 2019, Long Beach, CA, USA, June 16-20, 2019}}.
  \bibinfo{publisher}{Computer Vision Foundation / {IEEE}},
  \bibinfo{pages}{6988--6996}.
\newblock
\urldef\tempurl%
\url{https://doi.org/10.1109/CVPR.2019.00715}
\showDOI{\tempurl}


\bibitem[\protect\citeauthoryear{Wallace, Feng, Kandpal, Gardner, and
  Singh}{Wallace et~al\mbox{.}}{2019}]%
        {wallace-etal-2019-universal}
\bibfield{author}{\bibinfo{person}{Eric Wallace}, \bibinfo{person}{Shi Feng},
  \bibinfo{person}{Nikhil Kandpal}, \bibinfo{person}{Matt Gardner}, {and}
  \bibinfo{person}{Sameer Singh}.} \bibinfo{year}{2019}\natexlab{}.
\newblock \showarticletitle{Universal Adversarial Triggers for Attacking and
  Analyzing {NLP}}. In \bibinfo{booktitle}{\emph{Proceedings of the 2019
  Conference on Empirical Methods in Natural Language Processing and the 9th
  International Joint Conference on Natural Language Processing
  (EMNLP-IJCNLP)}}. \bibinfo{publisher}{Association for Computational
  Linguistics}, \bibinfo{address}{Hong Kong, China},
  \bibinfo{pages}{2153--2162}.
\newblock
\urldef\tempurl%
\url{https://doi.org/10.18653/v1/D19-1221}
\showDOI{\tempurl}


\bibitem[\protect\citeauthoryear{Wang, Wen, Chen, Huang, Zhang, and Zheng}{Wang
  et~al\mbox{.}}{2020}]%
        {DBLP:journals/corr/abs-2009-09841}
\bibfield{author}{\bibinfo{person}{Zifeng Wang}, \bibinfo{person}{Rui Wen},
  \bibinfo{person}{Xi Chen}, \bibinfo{person}{Shao{-}Lun Huang},
  \bibinfo{person}{Ningyu Zhang}, {and} \bibinfo{person}{Yefeng Zheng}.}
  \bibinfo{year}{2020}\natexlab{}.
\newblock \showarticletitle{Finding Influential Instances for Distantly
  Supervised Relation Extraction}.
\newblock \bibinfo{journal}{\emph{CoRR}}  \bibinfo{volume}{abs/2009.09841}
  (\bibinfo{year}{2020}).
\newblock
\showeprint[arxiv]{2009.09841}
\urldef\tempurl%
\url{https://arxiv.org/abs/2009.09841}
\showURL{%
\tempurl}


\bibitem[\protect\citeauthoryear{Wiegreffe and Pinter}{Wiegreffe and
  Pinter}{2019}]%
        {DBLP:conf/emnlp/WiegreffeP19}
\bibfield{author}{\bibinfo{person}{Sarah Wiegreffe} {and}
  \bibinfo{person}{Yuval Pinter}.} \bibinfo{year}{2019}\natexlab{}.
\newblock \showarticletitle{Attention is not not Explanation}. In
  \bibinfo{booktitle}{\emph{Proceedings of the 2019 Conference on Empirical
  Methods in Natural Language Processing and the 9th International Joint
  Conference on Natural Language Processing, {EMNLP-IJCNLP} 2019, Hong Kong,
  China, November 3-7, 2019}}, \bibfield{editor}{\bibinfo{person}{Kentaro
  Inui}, \bibinfo{person}{Jing Jiang}, \bibinfo{person}{Vincent Ng}, {and}
  \bibinfo{person}{Xiaojun Wan}} (Eds.). \bibinfo{publisher}{Association for
  Computational Linguistics}, \bibinfo{pages}{11--20}.
\newblock
\urldef\tempurl%
\url{https://doi.org/10.18653/v1/D19-1002}
\showDOI{\tempurl}


\bibitem[\protect\citeauthoryear{Yang, Dai, Yang, Carbonell, Salakhutdinov, and
  Le}{Yang et~al\mbox{.}}{2019}]%
        {DBLP:conf/nips/YangDYCSL19}
\bibfield{author}{\bibinfo{person}{Zhilin Yang}, \bibinfo{person}{Zihang Dai},
  \bibinfo{person}{Yiming Yang}, \bibinfo{person}{Jaime~G. Carbonell},
  \bibinfo{person}{Ruslan Salakhutdinov}, {and} \bibinfo{person}{Quoc~V. Le}.}
  \bibinfo{year}{2019}\natexlab{}.
\newblock \showarticletitle{XLNet: Generalized Autoregressive Pretraining for
  Language Understanding}. In \bibinfo{booktitle}{\emph{Advances in Neural
  Information Processing Systems 32: Annual Conference on Neural Information
  Processing Systems 2019, NeurIPS 2019, December 8-14, 2019, Vancouver, BC,
  Canada}}, \bibfield{editor}{\bibinfo{person}{Hanna~M. Wallach},
  \bibinfo{person}{Hugo Larochelle}, \bibinfo{person}{Alina Beygelzimer},
  \bibinfo{person}{Florence d'Alch{\'{e}}{-}Buc}, \bibinfo{person}{Emily~B.
  Fox}, {and} \bibinfo{person}{Roman Garnett}} (Eds.).
  \bibinfo{pages}{5754--5764}.
\newblock
\urldef\tempurl%
\url{https://proceedings.neurips.cc/paper/2019/hash/dc6a7e655d7e5840e66733e9ee67cc69-Abstract.html}
\showURL{%
\tempurl}


\bibitem[\protect\citeauthoryear{Ye, Zhang, Deng, Chen, Tan, Huang, and
  Chen}{Ye et~al\mbox{.}}{2020}]%
        {DBLP:journals/corr/abs-2009-06207}
\bibfield{author}{\bibinfo{person}{Hongbin Ye}, \bibinfo{person}{Ningyu Zhang},
  \bibinfo{person}{Shumin Deng}, \bibinfo{person}{Mosha Chen},
  \bibinfo{person}{Chuanqi Tan}, \bibinfo{person}{Fei Huang}, {and}
  \bibinfo{person}{Huajun Chen}.} \bibinfo{year}{2020}\natexlab{}.
\newblock \showarticletitle{Contrastive Triple Extraction with Generative
  Transformer}.
\newblock \bibinfo{journal}{\emph{CoRR}}  \bibinfo{volume}{abs/2009.06207}
  (\bibinfo{year}{2020}).
\newblock
\showeprint[arxiv]{2009.06207}
\urldef\tempurl%
\url{https://arxiv.org/abs/2009.06207}
\showURL{%
\tempurl}


\bibitem[\protect\citeauthoryear{Yu, Zhang, Deng, Ye, Zhang, and Chen}{Yu
  et~al\mbox{.}}{2020a}]%
        {DBLP:conf/coling/YuZDYZC20}
\bibfield{author}{\bibinfo{person}{Haiyang Yu}, \bibinfo{person}{Ningyu Zhang},
  \bibinfo{person}{Shumin Deng}, \bibinfo{person}{Hongbin Ye},
  \bibinfo{person}{Wei Zhang}, {and} \bibinfo{person}{Huajun Chen}.}
  \bibinfo{year}{2020}\natexlab{a}.
\newblock \showarticletitle{Bridging Text and Knowledge with Multi-Prototype
  Embedding for Few-Shot Relational Triple Extraction}. In
  \bibinfo{booktitle}{\emph{Proceedings of the 28th International Conference on
  Computational Linguistics, {COLING} 2020, Barcelona, Spain (Online), December
  8-13, 2020}}, \bibfield{editor}{\bibinfo{person}{Donia Scott},
  \bibinfo{person}{N{\'{u}}ria Bel}, {and} \bibinfo{person}{Chengqing Zong}}
  (Eds.). \bibinfo{publisher}{International Committee on Computational
  Linguistics}, \bibinfo{pages}{6399--6410}.
\newblock
\urldef\tempurl%
\url{https://doi.org/10.18653/v1/2020.coling-main.563}
\showDOI{\tempurl}


\bibitem[\protect\citeauthoryear{Yu, Zhang, Deng, Yuan, Jia, and Chen}{Yu
  et~al\mbox{.}}{2020b}]%
        {DBLP:journals/corr/abs-2009-07022}
\bibfield{author}{\bibinfo{person}{Haiyang Yu}, \bibinfo{person}{Ningyu Zhang},
  \bibinfo{person}{Shumin Deng}, \bibinfo{person}{Zonggang Yuan},
  \bibinfo{person}{Yantao Jia}, {and} \bibinfo{person}{Huajun Chen}.}
  \bibinfo{year}{2020}\natexlab{b}.
\newblock \showarticletitle{The Devil is the Classifier: Investigating Long
  Tail Relation Classification with Decoupling Analysis}.
\newblock \bibinfo{journal}{\emph{CoRR}}  \bibinfo{volume}{abs/2009.07022}
  (\bibinfo{year}{2020}).
\newblock
\showeprint[arxiv]{2009.07022}
\urldef\tempurl%
\url{https://arxiv.org/abs/2009.07022}
\showURL{%
\tempurl}


\bibitem[\protect\citeauthoryear{Zang, Qi, Yang, Liu, Zhang, Liu, and Sun}{Zang
  et~al\mbox{.}}{2020}]%
        {zang-etal-2020-word}
\bibfield{author}{\bibinfo{person}{Yuan Zang}, \bibinfo{person}{Fanchao Qi},
  \bibinfo{person}{Chenghao Yang}, \bibinfo{person}{Zhiyuan Liu},
  \bibinfo{person}{Meng Zhang}, \bibinfo{person}{Qun Liu}, {and}
  \bibinfo{person}{Maosong Sun}.} \bibinfo{year}{2020}\natexlab{}.
\newblock \showarticletitle{Word-level Textual Adversarial Attacking as
  Combinatorial Optimization}. In \bibinfo{booktitle}{\emph{Proceedings of the
  58th Annual Meeting of the Association for Computational Linguistics}}.
  \bibinfo{publisher}{Association for Computational Linguistics},
  \bibinfo{address}{Online}, \bibinfo{pages}{6066--6080}.
\newblock
\urldef\tempurl%
\url{https://doi.org/10.18653/v1/2020.acl-main.540}
\showDOI{\tempurl}


\bibitem[\protect\citeauthoryear{Zeng, Liu, Lai, Zhou, and Zhao}{Zeng
  et~al\mbox{.}}{2014}]%
        {DBLP:conf/coling/ZengLLZZ14}
\bibfield{author}{\bibinfo{person}{Daojian Zeng}, \bibinfo{person}{Kang Liu},
  \bibinfo{person}{Siwei Lai}, \bibinfo{person}{Guangyou Zhou}, {and}
  \bibinfo{person}{Jun Zhao}.} \bibinfo{year}{2014}\natexlab{}.
\newblock \showarticletitle{Relation Classification via Convolutional Deep
  Neural Network}. In \bibinfo{booktitle}{\emph{{COLING} 2014, 25th
  International Conference on Computational Linguistics, Proceedings of the
  Conference: Technical Papers, August 23-29, 2014, Dublin, Ireland}},
  \bibfield{editor}{\bibinfo{person}{Jan Hajic} {and} \bibinfo{person}{Junichi
  Tsujii}} (Eds.). \bibinfo{publisher}{{ACL}}, \bibinfo{pages}{2335--2344}.
\newblock
\urldef\tempurl%
\url{https://www.aclweb.org/anthology/C14-1220/}
\showURL{%
\tempurl}


\bibitem[\protect\citeauthoryear{Zhang, Deng, Bi, Yu, Yang, Chen, Huang, Zhang,
  and Chen}{Zhang et~al\mbox{.}}{2020a}]%
        {DBLP:conf/emnlp/ZhangDBYYCHZC20}
\bibfield{author}{\bibinfo{person}{Ningyu Zhang}, \bibinfo{person}{Shumin
  Deng}, \bibinfo{person}{Zhen Bi}, \bibinfo{person}{Haiyang Yu},
  \bibinfo{person}{Jiacheng Yang}, \bibinfo{person}{Mosha Chen},
  \bibinfo{person}{Fei Huang}, \bibinfo{person}{Wei Zhang}, {and}
  \bibinfo{person}{Huajun Chen}.} \bibinfo{year}{2020}\natexlab{a}.
\newblock \showarticletitle{OpenUE: An Open Toolkit of Universal Extraction
  from Text}. In \bibinfo{booktitle}{\emph{Proceedings of the 2020 Conference
  on Empirical Methods in Natural Language Processing: System Demonstrations,
  {EMNLP} 2020 - Demos, Online, November 16-20, 2020}},
  \bibfield{editor}{\bibinfo{person}{Qun Liu} {and} \bibinfo{person}{David
  Schlangen}} (Eds.). \bibinfo{publisher}{Association for Computational
  Linguistics}, \bibinfo{pages}{1--8}.
\newblock
\urldef\tempurl%
\url{https://doi.org/10.18653/v1/2020.emnlp-demos.1}
\showDOI{\tempurl}


\bibitem[\protect\citeauthoryear{Zhang, Deng, Li, Chen, Zhang, and Chen}{Zhang
  et~al\mbox{.}}{2020b}]%
        {DBLP:conf/emnlp/ZhangDLCZC20}
\bibfield{author}{\bibinfo{person}{Ningyu Zhang}, \bibinfo{person}{Shumin
  Deng}, \bibinfo{person}{Juan Li}, \bibinfo{person}{Xi Chen},
  \bibinfo{person}{Wei Zhang}, {and} \bibinfo{person}{Huajun Chen}.}
  \bibinfo{year}{2020}\natexlab{b}.
\newblock \showarticletitle{Summarizing Chinese Medical Answer with Graph
  Convolution Networks and Question-focused Dual Attention}. In
  \bibinfo{booktitle}{\emph{Proceedings of the 2020 Conference on Empirical
  Methods in Natural Language Processing: Findings, {EMNLP} 2020, Online Event,
  16-20 November 2020}}, \bibfield{editor}{\bibinfo{person}{Trevor Cohn},
  \bibinfo{person}{Yulan He}, {and} \bibinfo{person}{Yang Liu}} (Eds.).
  \bibinfo{publisher}{Association for Computational Linguistics},
  \bibinfo{pages}{15--24}.
\newblock
\urldef\tempurl%
\url{https://doi.org/10.18653/v1/2020.findings-emnlp.2}
\showDOI{\tempurl}


\bibitem[\protect\citeauthoryear{Zhang, Deng, Sun, Chen, Zhang, and Chen}{Zhang
  et~al\mbox{.}}{2020c}]%
        {DBLP:conf/www/ZhangDSCZC20}
\bibfield{author}{\bibinfo{person}{Ningyu Zhang}, \bibinfo{person}{Shumin
  Deng}, \bibinfo{person}{Zhanlin Sun}, \bibinfo{person}{Jiaoyan Chen},
  \bibinfo{person}{Wei Zhang}, {and} \bibinfo{person}{Huajun Chen}.}
  \bibinfo{year}{2020}\natexlab{c}.
\newblock \showarticletitle{Relation Adversarial Network for Low Resource
  Knowledge Graph Completion}. In \bibinfo{booktitle}{\emph{{WWW} '20: The Web
  Conference 2020, Taipei, Taiwan, April 20-24, 2020}},
  \bibfield{editor}{\bibinfo{person}{Yennun Huang}, \bibinfo{person}{Irwin
  King}, \bibinfo{person}{Tie{-}Yan Liu}, {and} \bibinfo{person}{Maarten van
  Steen}} (Eds.). \bibinfo{publisher}{{ACM} / {IW3C2}}, \bibinfo{pages}{1--12}.
\newblock
\urldef\tempurl%
\url{https://doi.org/10.1145/3366423.3380089}
\showDOI{\tempurl}


\bibitem[\protect\citeauthoryear{Zhang, Deng, Sun, Chen, Zhang, and Chen}{Zhang
  et~al\mbox{.}}{2018}]%
        {zhang-etal-2018-attention}
\bibfield{author}{\bibinfo{person}{Ningyu Zhang}, \bibinfo{person}{Shumin
  Deng}, \bibinfo{person}{Zhanling Sun}, \bibinfo{person}{Xi Chen},
  \bibinfo{person}{Wei Zhang}, {and} \bibinfo{person}{Huajun Chen}.}
  \bibinfo{year}{2018}\natexlab{}.
\newblock \showarticletitle{Attention-Based Capsule Networks with Dynamic
  Routing for Relation Extraction}. In \bibinfo{booktitle}{\emph{Proceedings of
  the 2018 Conference on Empirical Methods in Natural Language Processing}}.
  \bibinfo{publisher}{Association for Computational Linguistics},
  \bibinfo{address}{Brussels, Belgium}, \bibinfo{pages}{986--992}.
\newblock
\urldef\tempurl%
\url{https://doi.org/10.18653/v1/D18-1120}
\showDOI{\tempurl}


\bibitem[\protect\citeauthoryear{Zhang, Deng, Sun, Wang, Chen, Zhang, and
  Chen}{Zhang et~al\mbox{.}}{2019}]%
        {DBLP:conf/naacl/ZhangDSWCZC19}
\bibfield{author}{\bibinfo{person}{Ningyu Zhang}, \bibinfo{person}{Shumin
  Deng}, \bibinfo{person}{Zhanlin Sun}, \bibinfo{person}{Guanying Wang},
  \bibinfo{person}{Xi Chen}, \bibinfo{person}{Wei Zhang}, {and}
  \bibinfo{person}{Huajun Chen}.} \bibinfo{year}{2019}\natexlab{}.
\newblock \showarticletitle{Long-tail Relation Extraction via Knowledge Graph
  Embeddings and Graph Convolution Networks}. In
  \bibinfo{booktitle}{\emph{Proceedings of the 2019 Conference of the North
  American Chapter of the Association for Computational Linguistics: Human
  Language Technologies, {NAACL-HLT} 2019, Minneapolis, MN, USA, June 2-7,
  2019, Volume 1 (Long and Short Papers)}},
  \bibfield{editor}{\bibinfo{person}{Jill Burstein}, \bibinfo{person}{Christy
  Doran}, {and} \bibinfo{person}{Thamar Solorio}} (Eds.).
  \bibinfo{publisher}{Association for Computational Linguistics},
  \bibinfo{pages}{3016--3025}.
\newblock
\urldef\tempurl%
\url{https://doi.org/10.18653/v1/n19-1306}
\showDOI{\tempurl}


\bibitem[\protect\citeauthoryear{Zhang, Jia, Yin, Dong, Gao, and Hua}{Zhang
  et~al\mbox{.}}{2020d}]%
        {DBLP:journals/corr/abs-2008-10813}
\bibfield{author}{\bibinfo{person}{Ningyu Zhang}, \bibinfo{person}{Qianghuai
  Jia}, \bibinfo{person}{Kangping Yin}, \bibinfo{person}{Liang Dong},
  \bibinfo{person}{Feng Gao}, {and} \bibinfo{person}{Nengwei Hua}.}
  \bibinfo{year}{2020}\natexlab{d}.
\newblock \showarticletitle{Conceptualized Representation Learning for Chinese
  Biomedical Text Mining}.
\newblock \bibinfo{journal}{\emph{CoRR}}  \bibinfo{volume}{abs/2008.10813}
  (\bibinfo{year}{2020}).
\newblock
\showeprint[arxiv]{2008.10813}
\urldef\tempurl%
\url{https://arxiv.org/abs/2008.10813}
\showURL{%
\tempurl}


\bibitem[\protect\citeauthoryear{Zhang, Li, Deng, Yu, Cheng, Zhang, and
  Chen}{Zhang et~al\mbox{.}}{2020e}]%
        {DBLP:journals/corr/abs-2009-06206}
\bibfield{author}{\bibinfo{person}{Ningyu Zhang}, \bibinfo{person}{Luoqiu Li},
  \bibinfo{person}{Shumin Deng}, \bibinfo{person}{Haiyang Yu},
  \bibinfo{person}{Xu Cheng}, \bibinfo{person}{Wei Zhang}, {and}
  \bibinfo{person}{Huajun Chen}.} \bibinfo{year}{2020}\natexlab{e}.
\newblock \showarticletitle{Can Fine-tuning Pre-trained Models Lead to Perfect
  NLP? {A} Study of the Generalizability of Relation Extraction}.
\newblock \bibinfo{journal}{\emph{CoRR}}  \bibinfo{volume}{abs/2009.06206}
  (\bibinfo{year}{2020}).
\newblock
\showeprint[arxiv]{2009.06206}
\urldef\tempurl%
\url{https://arxiv.org/abs/2009.06206}
\showURL{%
\tempurl}


\bibitem[\protect\citeauthoryear{Zhang, Sheng, Alhazmi, and Li}{Zhang
  et~al\mbox{.}}{2020f}]%
        {DBLP:journals/tist/ZhangSAL20}
\bibfield{author}{\bibinfo{person}{Wei~Emma Zhang}, \bibinfo{person}{Quan~Z.
  Sheng}, \bibinfo{person}{Ahoud Abdulrahmn~F. Alhazmi}, {and}
  \bibinfo{person}{Chenliang Li}.} \bibinfo{year}{2020}\natexlab{f}.
\newblock \showarticletitle{Adversarial Attacks on Deep-learning Models in
  Natural Language Processing: {A} Survey}.
\newblock \bibinfo{journal}{\emph{{ACM} Trans. Intell. Syst. Technol.}}
  \bibinfo{volume}{11}, \bibinfo{number}{3} (\bibinfo{year}{2020}),
  \bibinfo{pages}{24:1--24:41}.
\newblock
\urldef\tempurl%
\url{https://doi.org/10.1145/3374217}
\showDOI{\tempurl}


\bibitem[\protect\citeauthoryear{Zhang, Zhong, Chen, Angeli, and Manning}{Zhang
  et~al\mbox{.}}{2017}]%
        {DBLP:conf/emnlp/ZhangZCAM17}
\bibfield{author}{\bibinfo{person}{Yuhao Zhang}, \bibinfo{person}{Victor
  Zhong}, \bibinfo{person}{Danqi Chen}, \bibinfo{person}{Gabor Angeli}, {and}
  \bibinfo{person}{Christopher~D. Manning}.} \bibinfo{year}{2017}\natexlab{}.
\newblock \showarticletitle{Position-aware Attention and Supervised Data
  Improve Slot Filling}. In \bibinfo{booktitle}{\emph{Proceedings of the 2017
  Conference on Empirical Methods in Natural Language Processing, {EMNLP} 2017,
  Copenhagen, Denmark, September 9-11, 2017}},
  \bibfield{editor}{\bibinfo{person}{Martha Palmer}, \bibinfo{person}{Rebecca
  Hwa}, {and} \bibinfo{person}{Sebastian Riedel}} (Eds.).
  \bibinfo{publisher}{Association for Computational Linguistics},
  \bibinfo{pages}{35--45}.
\newblock
\urldef\tempurl%
\url{https://doi.org/10.18653/v1/d17-1004}
\showDOI{\tempurl}


\end{thebibliography}

\end{document}